\documentstyle{nemlap}
\input{psfig}
\begin{document}
\title{The Proper Treatment of Optimality\\
in Computational Phonology}
\author{Lauri Karttunen}
\institute{Xerox Research Centre Europe\\
6, chemin de Maupertuis\\
38240 Meylan, France}
\maketitle

\abstract{This paper presents a novel formalization of optimality
theory. Unlike previous treatments of optimality in computational
linguistics, starting with Ellison (1994), the new approach does not
require any explicit marking and counting of constraint violations.  It
is based on the notion of ``lenient composition'', defined as the
combination of {\em ordinary composition} and {\em priority union}.  If
an underlying form has outputs that can meet a given constraint, lenient
composition enforces the constraint; if none of the output candidates
meets the constraint, lenient composition allows all of them. For the
sake of greater efficiency, we may "leniently compose" the {\sc gen}
relation and all the constraints into a single finite-state transducer
that maps each underlying form directly into its optimal surface
realizations, and vice versa. Seen from this perspective, optimality
theory is surprisingly similar to the two older strains of finite-state
phonology: classical rewrite systems and two-level models. In
particular, the ranking of optimality constraints corresponds to the
ordering of rewrite rules.}

\section{Introduction}

It has been recognized for some time that Optimality Theory ({\sc ot}),
introduced by Prince and Smolensky \cite{prince+smolensky1993}, is
from a computational point of view closely related to classical
phonological rewrite systems (Chomsky and Halle
\cite{chomsky+halle1968}) and to two-level descriptions
(Koskenniemi \cite{koskenniemi1983}).

Ellison \cite{ellison1994a} observes that the {\sc gen} function of
{\sc ot} can be regarded as a regular relation and that {\sc ot}
constraints seem to be regular. Thus each constraint can be modeled as a
transducer that maps a string to a sequence of marks indicating the
presence or absence of a violation. The most optimal solution can then
be found by sorting and comparing the marks. Frank and Satta
\cite{frank+satta1998} give a formal proof that {\sc ot} models
can be construed as regular relations provided that the number of
violations is bounded. Eisner
\cite{eisner1997a,eisner1997b,eisner1997c} develops a typology
of {\sc ot} constraints that corresponds to two types of rules in
two-level descriptions: restrictions and prohibitions.

The practice of marking and counting constraint violations is closely
related to the tableau method introduced in Prince and Smolensky for
selecting the most optimal output candidate. Much of the current work in
optimality theory consists of constructing tableaux that demonstrate
the need for particular constraints and rankings that allow the
favored candidate to emerge with the best score.

From a computational viewpoint, this evaluation method is suboptimal.
Although the work of {\sc gen} and the assignment of violation marks can
be carried out by finite-state transducers, the sorting and counting of
the marks envisioned by Ellison and subsequent work (Walther
\cite{walther1996}) is an off-line activity that is not a finite-state
process. This kind of optimality computation cannot be
straightforwardly integrated with other types of linguistic processing
(morphological analysis, text-to-speech generation etc.) that are
commonly performed by means of finite-state transduction.

This paper demonstrates that the computation of the most optimal surface
realizations of any input string can be carried out entirely within a
finite-state calculus, subject to the limitation (Frank and Satta
\cite{frank+satta1998}) that the maximal number of violations that need
to be considered is bounded. We will show that optimality constraints
can be treated computationally in a similar manner to two-level
constraints and rewrite rules. For example, optimality constraints can
be merged with one another, respecting their ranking,
just as it is possible to merge rewrite rules and two-level
constraints. A system of optimality constraints can be imposed on a
finite-state lexicon creating a transducer that maps each member of a
possibly infinite set of lexical forms into its most optimal surface
realization, and vice versa.

For the sake of conciseness, we limit the discussion to optimality
theory as originally presented in Prince and Smolensky
\cite{prince+smolensky1993}. The techniques described below can
also be applied to the correspondence version of the theory (McCarthy
and Prince \cite{mccarthy+prince1995}) that expands the model to
encompass output/output constraints between reduplicant and base forms.

To set the stage for discussing the application and merging of
optimality constraints it is useful to look first at the corresponding
operations in the context of rewrite rules and two-level
constraints. Thus we can see both the similarities and the differences
among the three approaches.

\section{Background: rewrite rules and two-level constraints}

As is well-known, phonological rewrite rules and two-level constraints
can be implemented as finite-state transducers
(Johnson \cite{johnson1972}, Karttunen, Koskenniemi and Kaplan
\cite{karttunen+koskenniemi+kaplan1987}, Kaplan and Kay
\cite{kaplan+kay1994}).

The application of a system of rewrite rules to an input string can be
modeled as a cascade of transductions, that is, a sequence of
compositions that yields a relation mapping the input string to one or
more surface realizations. The application of a set of two-level
constraints is a combination of intersection and composition
(Karttunen \cite{karttunen1994}).

To illustrate the idea of rule application as composition, let us take
a concrete example, the well-known vowel alternations in Yokuts
(Kisseberth \cite{kisseberth1969}, Cole and
Kisseberth \cite{cole+kisseberth1995}, McCarthy \cite{mccarthy1998}).
Yokuts vowels are subject to three types of alternations:

\begin{itemize}
\item{Underspecified suffix vowels are rounded in the presence of
a stem vowel of the same height: {\em dub+hIn} $\rightarrow$ {\em
dubhun}, {\em bok'+Al} $\rightarrow$ {\em bok'ol}.}
\item{Long high vowels are lowered: {\em ?u:t+It} $\rightarrow$ {\em
?o:tut}, {\em mi:k+It} $\rightarrow$ {\em me:kit}.}
\item{Vowels are shortened in closed syllables: {\em sa:p}
$\rightarrow$ {\em sap}, {\em go:b+hIn} $\rightarrow$ {\em gobhin}.}
\end{itemize}

Because of examples such as {\em ?u:t+hIn} $\rightarrow$ {\em ?othun},
the rules must be applied in the given order. Rounding must precede
lowering because the suffix vowel in {\em ?u:t+hIn} emerges as {\em u}.
Shortening must follow lowering because the stem vowel in {\em ?u:t+hIn}
would otherwise remain high giving {\em ?uthun} rather than {\em ?othun}
as the final output.

These three rewrite rules can be formalized straightforwardly as
regular replace expressions (Karttunen \cite{karttunen1995}) and
compiled into finite-state transducers. The derivation {\em ?u:t+hIn}
$\rightarrow$ {\em ?othun} can thus be modeled as a cascade of three
compositions that yield a transducer that relates the input directly to
the final output.

\begin{figure}[here]
\vspace*{-10mm}
\begin{center}
  \centerline{\psfig{file=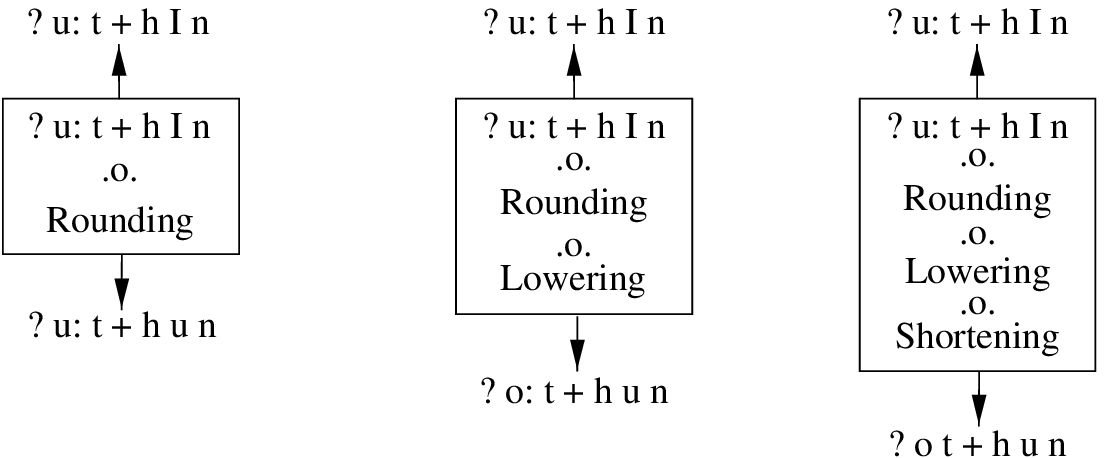}}
\vspace*{-2mm}
\caption{\label{cascade1} Cascade of rewrite rule applications.}
\end{center}
\vspace*{-15mm}
\end{figure}

The first step, the composition of the initial network (an identity
transducer on the string {\em ?u:t+hIn}) with the rounding transducer,
produces the network that maps between {\em ?u:t+hIn} and {\em
?u:t+hun}.  The symbol {\tt .o.} in Figure \ref{cascade1} denotes the
composition operation.

It is important to realize that the result of each rule application in
Figure \ref{cascade1} is not an output string but a relation.  The first
application produces a mapping from {\em ?u:t+hIn} to {\em ?u:t+hun}. In
essence, it is the original Rounding transducer restricted to the
specific input. The resulting network represents a relation between two
languages (= sets of strings). In this case both languages contain just
one string; but if the Rounding rule were optional, the output language
would contain two strings: one with, the other without rounding.

At the next step in Figure \ref{cascade1}, the intermediate output
created by the Rounding transducer is eliminated as a result of the
composition with the Lowering transducer. The final stage is a
transducer that maps directly between the input string and its surface
realization without any intermediate stages.

\begin{figure}[here]
\vspace*{-10mm}
\begin{center}
  \centerline{\psfig{file=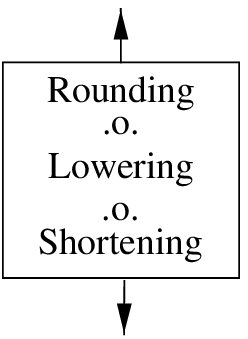}}
\caption{\label{yokuts} Yokuts vowel alternations.}
\end{center}
\vspace*{-15mm}
\end{figure}

We could achieve this same result in a different way: by first composing
the three rules to produce a transducer that maps any underlying form
directly to its Yokuts surface realization (Figure \ref{yokuts}) and then
applying the resulting single transducer to the particular input.

The small network (21 states) pictured in Figure \ref{yokuts} merges the
three rules and thus represents the complexity of Yokuts vowel
alternations without any ``serialism'', that is, without any
intermediate representations.

\begin{figure}[here]
\vspace*{-10mm}
\begin{center}
  \centerline{\psfig{file=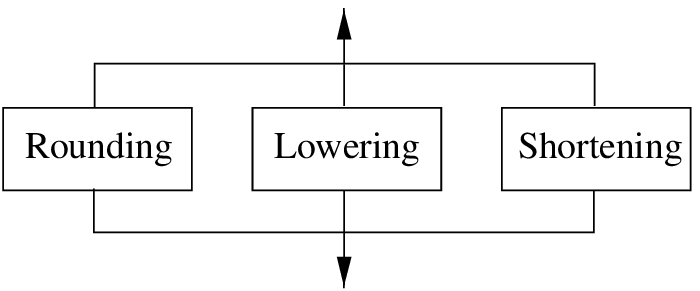}}
\caption{\label{2lyokuts} Parallel two-level constraints.}
\end{center}
\vspace*{-15mm}
\end{figure}

In the context of the two-level model, the Yokuts vowel alternations can
be described quite simply. The two-level version of the rounding rule
controls rounding by the lexical context. It ignores the surface
realization of the trigger, the underlyingly high stem vowel. The joint
effect of the lowering and shortening constraints is that a lexical {\em
u:} in {\em ?u:t+hIn} is realized as {\em o}. Thus a two-level
description of the Yokuts alternations consists of three rule
transducers operating in parallel (Figure \ref{2lyokuts}).

The application of a two-level system to an input can be formalized as
{\em intersecting composition} (Karttunen \cite{karttunen1994}). It
involves constructing a partial intersection of the constraint networks
and composing it with the input. We can of course carry out the
intersection of the rules independently of any particular input. This
merging operation results in the very same 21-state transducer as the
composition of the corresponding rewrite rules pictured in Figure
\ref{yokuts}.

Thus the two descriptions of Yokuts sketched above are completely
equivalent in that they yield the same mapping between underlying
and surface forms. They decompose the same complex vowel alternation
relation in different ways into a set of simpler relations that are
easily understood and manipulated.\footnote{For more discussion
of these issues, see Karttunen \cite{karttunen1993a}.} As we
will see shortly, optimality theory can be characterized as yet
another way of achieving this kind of decomposition.

The fundamental computational operation for rewrite rules is
composition, as it is involved both in the application of rules to
strings and in merging the rules themselves. For two-level rules, the
corresponding operations are intersecting composition and intersection.

Turning now to optimality theory, our main interest will be in finding
what the corresponding computations are in this new paradigm. What does
applying a constraint mean in the context of optimality theory?  Can
optimality constraints be merged while taking into account their
ranking?

\section{Optimality theory}

Optimality theory (Prince and Smolensky \cite{prince+smolensky1993})
abandons rewrite rules.  Rules are replaced by two new concepts: (1) a
universal function called {\sc gen} and (2) a set of ranked universal
constraints. {\sc gen} provides each input form with a (possibly
infinite) set of output candidates. The constraints eliminate all but
the best output candidate. Because many constraints are in conflict, it
may be impossible for any candidate to satisfy all of them. The winner
is determined by taking into consideration the language-specific ranking
of the constraints. The winning candidate is the one with the least
serious violations.

In order to explore the computational aspects of the theory it is useful
to focus on a concrete example, even simpler than the Yokuts vowel
alternation we just discussed.\footnote {The Yokuts case is problematic
for Optimality theory (Cole and Kisseberth \cite{cole+kisseberth1995},
McCarthy \cite{mccarthy1998}) because rounding depends on the height of
the stem vowel in the underlying representation. Cole and Kisseberth
offer a baroque version of the two-level solution. McCarthy strives
mightily to distinguish his ``sympathy'' candidates from the
intermediate representations postulated by the rewrite approach.} We
will take the familiar case of syllabification constraints discussed by
Prince and Smolensky \cite{prince+smolensky1993} and many subsequent
authors (Ellison \cite{ellison1994a}, Tesar \cite{tesar1995}, Hammond
\cite{hammond1997}).

\subsection{{\sc gen} for syllabification}

We assume that the input to {\sc gen} consists of strings of vowels {\tt
V} and consonants {\tt C}. {\sc gen} allows each segment to play a role
in the syllable or to remain ``unparsed''. A syllable contains at least
a nucleus and possibly an onset and a coda.

Let us assume that {\sc gen} marks these roles by inserting labeled
brackets around each input element. An input consonant such as {\em b}
will have three outputs {\tt 0[b]} (onset), {\tt D[b]} (coda), and {\tt
X[b]} (unparsed). Each vowel such as {\em a} will have two outputs, {\tt
N[a]} (nucleus) and {\tt X[a]} (unparsed), In addition, {\sc gen}
``overparses'', that is, it freely inserts empty onset {\tt O[~]},
nucleus {\tt N[~]}, and coda {\tt D[~]} brackets.

For the sake of concreteness, we give here an explicit definition of
{\sc gen} using the notation of the Xerox regular expression calculus
(Karttunen {\em et al} \cite{karttunen+others1996}). We define {\sc gen}
as the composition of four simple components, {\tt Input}, {\tt Parse},
{\tt OverParse}, and {\tt SyllableStructure}. The definitions of the
first three components are shown in Figure \ref{parse}.

\begin{figure}
\vspace*{-10mm}
\begin{verbatim}
                define Input     [C | V]*                           ;

                define Parse     C -> ["O[" | "D[" | "X["] ... "]"
                                           .o.
                                 V -> ["N[" | "X["] ... "]"         ;

                define OverParse [. .] (->) ["O["|"N["|"D["] "]"    ;
\end{verbatim}
\vspace*{-8mm}
\caption{\label{parse} Input, Parse, and OverParse}
\vspace*{-4mm}
\end{figure}

A replace expression of the type {\tt A -> B ... C} in the Xerox
calculus denotes a relation that wraps the prefix strings in {\tt B} and
the suffix strings in {\tt C} around every string in {\tt A}. Thus {\tt
Parse} is a transducer that inserts appropriate bracket pairs around
input segments. Consonants can be onsets, codas, or be ignored. Vowels
can be nuclei or be ignored. {\tt OverParse} inserts optionally unfilled
onsets, codas, and nuclei. The dotted brackets {\tt [. .]} specify that
only a single instance of a given bracket pair is inserted at any
position.

The role of the third {\sc gen} component, {\tt SyllableStructure}, is
to constrain the output of {\tt Parse} and {\tt OverParse}. A syllable
needs a nucleus, onsets and codas are optional; they must be in the
right order; unparsed elements may occur freely. For the sake of
clarity, we define {\tt SyllableStructure} with the help of four
auxiliary terms (Figure \ref{syllable}).

\begin{figure}[here]
\vspace*{-10mm}
\begin{verbatim}
                define Onset             "O[" (C) "]"   ;
                define Nucleus           "N[" (V) "]"   ;
                define Coda              "D[" (C) "]"   ;
                define Unparsed          "X[" [C|V] "]" ;

                define SyllableStructure [[(Onset) Nucleus (Coda)]/Unparsed]* ;
\end{verbatim}
\vspace*{-8mm}
\caption{\label{syllable} SyllableStructure}
\vspace*{-4mm}
\end{figure}

Round parentheses in the Xerox regular expression notation indicate
optionality. Thus {\tt (C)} in the definition of {\tt Onset} indicates
that onsets may be empty or filled with a consonant. Similarly, {\tt
(Onset)} in the definition of {\tt SyllableStructure} means that a
syllable may have or not have an onset. The effect of the {\tt /}
operator is to allow unparsed consonants and vowels to occur freely
within a syllable. The disjunction {\tt [C|V]} in the definition of {\tt
Unparsed} allows consonants and vowels to remain unparsed.

With these preliminaries we can now define {\sc gen} as a simple
composition of the four components (Figure \ref{gen}).

\begin{figure}[here]
\vspace*{-10mm}
\begin{verbatim}
                       define GEN        Input
                                          .o.
                                       OverParse
                                          .o.
                                         Parse
                                          .o.
                                   SyllableStructure ;
\end{verbatim}
\vspace*{-8mm}
\caption{\label{gen} {\sc gen} for syllabification}
\vspace*{-4mm}
\end{figure}

With the appropriate definitions for {\tt C} (consonants) and {\tt V}
(vowels), the expression in Figure \ref{gen} yields a transducer with 22
states and 229 arcs.

It is not necessary to include {\tt Input} in the definition of {\sc
gen} but it has technically a beneficial effect. The constraints have less
work to do when it is made explicit that the auxiliary bracket
alphabet is not included in the input.

Because {\sc gen} over- and underparses with wild abandon, it produces
a large number of output candidates even for very short inputs. For
example, applying {\sc gen} to the string {\em a} yields a relation
with 14 strings on the output side (Figure \ref{a.o.gen}).

\begin{figure}[here]
\vspace*{-10mm}
\begin{verbatim}
                                         N[a]
                                         N[a]N[]
                                         N[a]D[]
                                      N[]N[a]
                                      N[]N[a]N[]
                                      N[]N[a]D[]
                                      N[]X[a]
                                      N[]X[a]N[]
                                      N[]X[a]D[]
                                      O[]N[a]
                                      O[]N[a]N[]
                                      O[]N[a]D[]
                                      O[]X[a]N[]
                                         X[a]N[]
\end{verbatim}
\vspace*{-8mm}
\caption{\label{a.o.gen} {\sc gen} applied to {\em a}}
\vspace*{-4mm}
\end{figure}

The number of output candidates for {\em abracadabra} is nearly 1.7
million, although the network representing the mapping has only 193
states. It is evident that working with finite-state tools has
a significant advantage over manual tableau methods.

\subsection{Syllabification constraints}

The syllabification constraints of Prince and Smolensky
\cite{prince+smolensky1993} can easily be expressed as regular
expressions in the Xerox calculus. Figure \ref{constraints} lists the
five constraints with their translations.

\begin{figure}[here]
\vspace*{-10mm}
\begin{tabbing}
~~~~~~~~~~~~~~~\= Syllables must have onsets.
	~~~~~~~~~~~~~\= {\tt define HaveOns \tt N[" => "O[" (C) "]" \_~;}\\
\\
	\> Syllables must not have codas.
	\>	{\tt define NoCoda~~~	 \verb+~$+"D[" ; }\\
\\
	\> Input segments must be parsed.
	\>	{\tt define Parse~~~~	 \verb+~$+"X[" ; }\\
\\
	\> A nucleus position must be filled.
	\>	 {\tt define FillNuc	 \verb+~$+[ "N[" "]" ] ; }\\
\\
	\> An onset position must be filled.
	\>	 {\tt define FillOns	 \verb+~$+[ "O[" "]" ] ; }
\end{tabbing}
\vspace*{-8mm}
\caption{\label{constraints} {Syllabification constraints}}
\vspace*{-4mm}
\end{figure}

The definition of the {\tt HaveOns} constraint uses the {\em
restriction} operator \verb+=>+. It requires that any occurrence of the
nucleus bracket, {\tt [N}, must be immediately preceded by a filled {\tt
O[C]} or unfilled {\tt 0[~]} onset. The definitions of the other four
constraints are composed of the negation \verb+~+ and the contains
operator \verb+$+. For example, the {\tt NoCoda} constraint,
\verb+~$"D["+, can be read as ``does not contain {\tt D[}''. The {\tt
FillNuc} and {\tt FillOns} constraints forbid empty nucleus {\tt N[~]}
and onset {\tt O[~]} brackets.

These constraints compile into very small networks, the largest one,
{\tt HaveOns}, contains four states. Each constraint network encodes
an infinite regular language. For example, the {\tt HaveOns} language
includes all strings of any length that contain no instances of {\tt N[}
at all and all strings of any length in which every instance of {\tt N[}
is immediately preceded by an onset.

The identity relations on these constraint languages can be thought of
as filters.  For example, the identity relation on {\tt HaveOns} maps
all {\tt HaveOns} strings into themselves and blocks on all other
strings. In the following section, we will in fact consistently treat
the constraint networks as representing identity relations.

\subsection{Constraint application}

Having defined {\sc gen} and the five syllabification constraints we
are now in a position to address the main issue: {\em how are optimality
constraints applied?}

\begin{figure}[here]
\vspace*{-10mm}
\begin{center}
  \centerline{\psfig{file=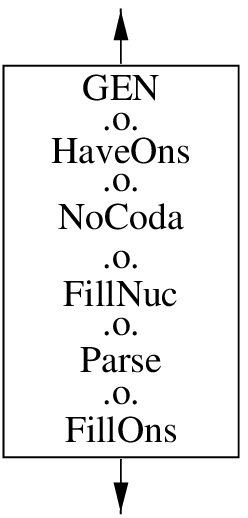}}
\vspace*{-4mm}
\caption{\label{merciless} Merciless cascade.}
\end{center}
\vspace*{-15mm}
\end{figure}

Given that {\sc gen} denotes a relation and that the constraints can be
thought of as identity relations on sets, the simplest idea is to
proceed in the same way as with the rewrite rules in Figure \ref{yokuts}.
We could compose {\sc gen} with the constraints to yield a transducer
that maps each input to its most optimal realization, letting the
ordering of the constraints in the cascade implement their ranking
(Figure \ref{merciless}).

But it is immediately obvious that composition does not work here as
intended. The 6-state transducer illustrated in Figure \ref{merciless}
works fine on inputs such as {\em panama} yielding {\tt
O[p]N[a]O[n]N[a]O[m]N[a]} but it fails to produce any output on inputs
like {\em america} that fail on some constraint. Only strings that have
a perfect output candidate survive this merciless cascade.  We need to
replace composition with some new operation to make this schema work
correctly.

\section{Lenient composition}

The necessary operation, let us call it {\em lenient composition}, is
not difficult to construct, but to our knowledge it has not previously
been defined. Frank and Satta \cite{frank+satta1998} come very close but
do not take the final step to encapsulate the notion. Hammond
\cite{hammond1997} has the idea but lacks the means to spell it out in
formal terms.

As the first step toward defining lenient composition, let us review an
old notion called {\em priority union} (Kaplan \cite{kaplan1987}).  This
term was originally defined as an operation for unifying two feature
structures in a way that eliminates any risk of failure by stipulating
that one of the two has priority in case of a
conflict.\footnote{\label{patr-ii}The {\sc dpatr} system at {\sc sri}
(Karttunen \cite{karttunen1986}) had the same operation with a less respectable
title.  It was called ``clobber''.} A finite-state version of this
notion has proved very useful in the management of transducer lexicons
(Kaplan and Newman \cite{kaplan+newman1997}).

\begin{figure}[here]
\vspace*{-15mm}
\begin{center}
  \centerline{\psfig{file=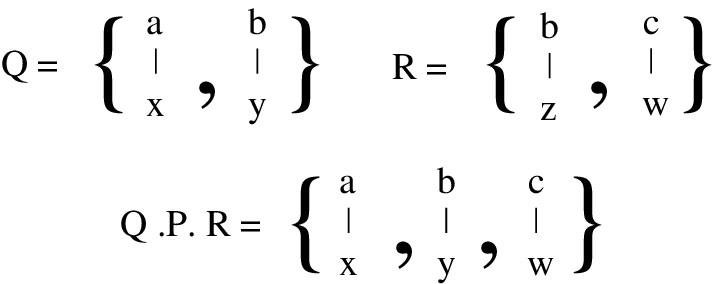}}
\vspace*{-2mm}
\caption{\label{priorityunion} Example of priority union.}
\end{center}
\vspace*{-15mm}
\end{figure}

Let us consider the relations {\tt Q} and {\tt R} depicted in Figure
\ref{priorityunion}. The {\tt Q} relation maps {\em a} to {\em x} and
{\em b} to {\em y}.  The {\tt R} relation maps {\em b} to {\em z} and
{\em c} to {\em w}. The priority union of {\tt Q} and {\tt R}, denoted
{\tt Q .P. R}, maps {\em a} to {\em x}, {\em b} to {\em y}, and {\em c}
to {\em w}. That is, it includes all the pairs from {\tt Q} and every
pair from R that has as its upper element a string that does not occur as
the upper string of any pair in {\tt Q}. If some string occurs as the
upper element of some pair in both {\tt Q} and {\tt R}, the priority
union of {\tt Q} and {\tt R} only includes the pair in {\tt
Q}. Consequently {\tt Q .P. R} in Figure \ref{priorityunion} maps {\em
b} to {\em y} instead of {\em z}.

The priority union operator {\tt .P.} can be defined in terms of other
regular expression operators in the Xerox calculus. A straightforward
definition is given in Figure \ref{Pdef}.

\begin{figure}[here]
\vspace*{-10mm}
\begin{verbatim}
                              Q .P. R = Q | [~[Q.u] .o. R]\end{verbatim}
\vspace*{-8mm}
\caption{\label{Pdef} Definition of priority union}
\vspace*{-4mm}
\end{figure}

The {\tt .u} operator in Figure \ref{Pdef} extracts the ``upper''
language from a regular relation. Thus the expression {\tt \verb+~+[Q.u]}
denotes the set of strings that do not occur on the upper side of
the {\tt Q} relation. The effect of the composition in
Figure \ref{Pdef} is to restrict R to mappings that concern
strings that are not mapped to anything in {\tt Q}. Only this subset
of R is unioned with {\tt Q}.

We define the desired operation, lenient composition, denoted {\tt .O.},
as a combination of ordinary composition and priority union (Figure
\ref{Odef}).

\begin{figure}[here]
\vspace*{-10mm}
\begin{verbatim}
                              R .O. C  = [R .o. C] .P. R \end{verbatim}
\vspace*{-8mm}
\caption{\label{Odef} Definition of lenient composition}
\vspace*{-4mm}
\end{figure}

To better visualize the effect of the operation defined in Figure
\ref{Odef} one may think of the relation {\tt R} as a set of mappings
induced by {\sc gen} and the relation {\tt C} as one of the constraints
defined in Figure \ref{constraints}. The left side of the priority
union, {\tt [R .o. C]} restricts R to mappings that satisfy the
constraint. That is, any pair whose lower side string is not in {\tt C}
will be eliminated. If some string in the upper language of {\tt R} has
no counterpart on the lower side that meets the constraint, then it is
not present in {\tt [R .o. C].u} but, for that very reason, it will be
``rescued'' by the priority union.  In other words, if an underlying
form has some output that can meet the given constraint, lenient
composition enforces the constraint.  If an underlying form has no
output candidates that meet the constraint, then the underlying form and
all its outputs are retained.  The definition of lenient composition
entails that the upper language of {\tt R} is preserved in {\tt R
.O. C}.

Many people, including Hammond \cite{hammond1997} and Frank and
Satta \cite{frank+satta1998}, have independently had a similar
idea without conceiving it as a finite-state operation.\footnote{Hammond
implements a pruning operation that removes output candidates under the
condition that ``pruning cannot reduce the candidate set to null'' (p
13). Frank and Satta (p. 7) describe a process of ``conditional
intersection'' that enforces a constraint if it can be met and does
nothing otherwise.} If one already knows about priority union, lenient
composition is an obvious idea.

\begin{figure}[here]
\vspace*{-15mm}
\begin{center}
  \centerline{\psfig{file=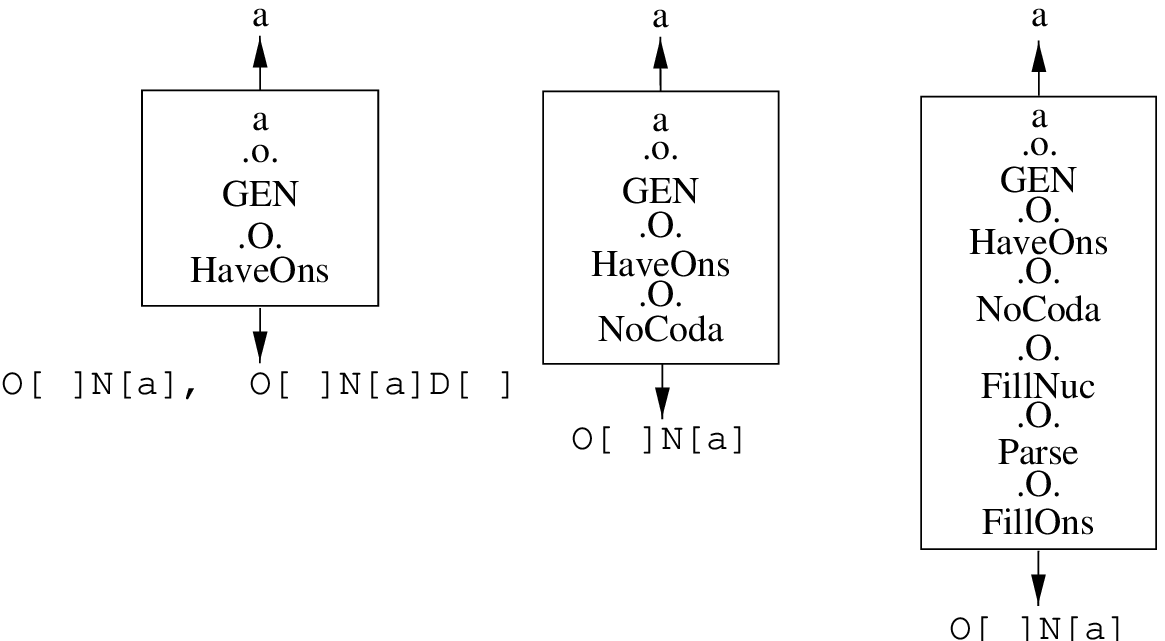}}
\vspace*{-2mm}
\caption{\label{cascade2} Cascade of constraint applications.}
\end{center}
\vspace*{-15mm}
\end{figure}

Let us illustrate the effect of lenient composition starting with the
example in Figure \ref{a.o.gen} The composition of the input {\em a}
with {\sc gen} yields a relation that maps {\em a} to the 14 outputs in
Figure \ref{a.o.gen}.  We will leniently compose this relation with each
of the constraints in the order of their ranking, starting with the {\tt
HaveOns} constraint (Figure \ref{cascade2}). The lower-case operator
{\tt .o.}  stands for ordinary composition, the upper case {\tt .O.} for
lenient composition.

As Figure \ref{cascade2} illustrates, applying {\tt HaveOns} by lenient
composition removes most of the 14 output candidates produced by {\sc
gen}. The resulting relation maps {\em a} to two outputs {\tt O[~]N[a]}
and {\tt O[~]N[a]D[~]}. The next highest-ranking constraint, {\tt
NoCoda}, removes the latter alternative. The twelve candidates that
were eliminated by the first lenient composition are no longer under
consideration.

The next two constraints in the sequence, {\tt FillNuc} and {\tt Parse},
obviously do not change the relation because the one remaining output
candidate, {\tt O[~]N[a]}, satisfies them. Up to this point, the
distinction between lenient and ordinary composition does not make any
difference because we have not exhausted the set of output
candidates. However, when we bring in the last constraint, {\tt
FillOns}, the right half of the definition in Figure \ref{Odef} has to
come to the rescue; otherwise there would be no output for {\em a}.

This example demonstrates that the application of optimality constraints
can be thought of as a cascade of lenient compositions that carry down
an ever decreasing number of output candidates without allowing the set
to become empty. Instead of {\em intermediate representations} (c.f.
Figure \ref{cascade1}) there are {\em intermediate candidate
populations} corresponding to the columns in the left-to-right ordering
of the constraint tableau.

\begin{figure}[here]
\vspace*{-10mm}
\begin{center}
  \centerline{\psfig{file=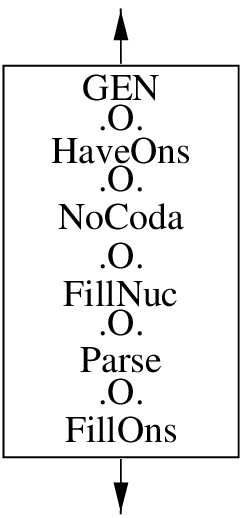}}
\vspace*{-4mm}
\caption{\label{lenient} Lenient cascade}
\end{center}
\vspace*{-15mm}
\end{figure}

Instead of applying the constraints one by one to the output
provided by {\sc gen} for a particular input, we may also leniently
compose the {\sc gen} relation itself with the constraints. Thus
the suggestion made in Figure \ref{merciless} is (nearly) correct
after all, provided that we replace ordinary composition with
lenient composition (Figure \ref{lenient}).

The composite single transducer shown in Figure \ref{lenient} maps
{\em a} and any other input directly into its viable outputs without
ever producing any failing candidates.

\section{Multiple violations}

However, we have not yet addressed one very important issue. It is not
sufficient to obey the ranking of the constraints. If two or more output
candidates violate the same constraint multiple times we should prefer
the candidate or candidates with the smallest number of violations.
This does not come for free. The system that we have sketched so far
does not make that distinction. If the input form has no perfect
outputs, we may get a set of outputs that differ with respect to the
number of constraint violations. For example, the transducer in Figure
\ref{lenient} gives three outputs for the string {\em bebop} (Figure
\ref{bebop}).

\begin{figure}[here]
\vspace*{-10mm}
\begin{verbatim}
                                  O[b]N[e]X[b]X[o]X[p]
                                  O[b]N[e]O[b]N[o]X[p]
                                  X[b]X[e]O[b]N[o]X[p]
\end{verbatim}
\vspace*{-8mm}
\caption{\label{bebop} Too many outputs}
\vspace*{-4mm}
\end{figure}

Because {\em bebop} has no output that meets the {\tt Parse} constraint,
lenient composition allows all outputs that contain a {\tt Parse}
violation regardless of the number of violations. Here the second
alternative with just one violation should win but it does not.

Instead of viewing {\tt Parse} as a single constraint, we need to
reconstruct it as a series of ever more relaxed parse constraints.
The \verb+^>n+ operator in Figure \ref{parseN} means ``more than
{\em n} iterations''.

\begin{figure}[here]
\vspace*{-10mm}
\begin{verbatim}
                                define Parse    ~$["X["]     ;
                                define Parse1  ~[[$"X["]^>1] ;
                                define Parse2  ~[[$"X["]^>2] ;
                                        ...
                                define ParseN  ~[[$"X["]^>N] ;\end{verbatim}
\vspace*{-8mm}
\caption{\label{parseN} A family of {\tt Parse} constraints}
\vspace*{-4mm}
\end{figure}

Our original {\tt Parse} constraint is violated by a single unparsed
element. {\tt Parse1} allows one unparsed element. {\tt
Parse2} allows up to two violations, and {\tt ParseN} up to {\em N}
violations.

The single {\tt Parse} line in Figure \ref{lenient} must be replaced by
the sequence of lenient compositions in Figure \ref{gradient} up to
some chosen {\em N}.

\begin{figure}[here]
\vspace*{-10mm}
\begin{center}
{\tt
Parse\\
.O.\\
Parse1\\
.O.\\
Parse2\\
.O.\\
ParseN}
\vspace*{-2mm}
\caption{\label{gradient} Gradient {\tt Parse} constraint}
\vspace*{-10mm}
\end{center}
\end{figure}

If an input string has at least one output form that meets the {\tt
Parse} constraint (no violations), all the competing output forms with
{\tt Parse} violations are eliminated. Failing that, if the input string
has at least one output form with just one violation, all the outputs
with more violations are eliminated. And so on.

The particular order in which the individual parse constraints apply
actually has no effect here on the final outcome because the constraint
languages are in a strict subset relation: {\tt Parse} $\subset$ {\tt
Parse1} $\subset$ {\tt Parse2} $\subset$ $\ldots$ {\tt
ParseN}.\footnote{Thanks to Jason Eisner (p.c.) for this observation.}
For example, if the best candidate incurs two violations, it is in {\tt
Parse2} and in all the weaker constraints. The ranking in Figure
\ref{gradient} determines only the order in which the losing candidates
are eliminated. If we start with the strictest constraint, all the
losers are eliminated at once when {\tt Parse2} is applied; if we start
with a weaker constraint, some output candidates will be eliminated
earlier than others but the winner remains the same.

As the number of constraints goes up, so does the size of the combined
constraint network in Figure \ref{lenient}, from 66 states (no {\tt
Parse} violations) to 248 (at most five violations). It maps {\em bebop}
to {\tt O[b]N[e]\-O[b]N[o]\-X[p]} and {\em abracadabra} to {\tt
O[]N[a]\-X[b]\-O[r]N[a]\-O[c]N[a]\-O[d]N[a]\-X[b]\-O[r]N[a]}
correctly and instantaneously.

It is immediately evident that while we can construct a cascade of
constraints that prefer {\em n} violations to {\em n+1} violations up to
any given {\em n}, there is no way in a finite-state system to express
the general idea that fewer violations is better than more violations.
As Frank and Satta \cite{frank+satta1998} point out, finite-state
constraints cannot make infinitely many distinctions of well-formedness.
It is not likely that this limitation is a serious obstacle to practical
optimality computations with finite-state systems as the number of
constraint violations that need to be taken into account is generally
small.

It is curious that violation counting should emerge as the crucial issue
that potentially pushes optimality theory out of the finite-state domain,
thus making it formally more powerful than rewrite systems and two-level
models. It has never been presented as an argument against the older
models that they do not allow unlimited counting. It is not clear
whether the additional power constitutes an asset or an embarrassment
for {\sc ot}.

\section{Conclusion}

This novel formalization of optimality theory has several technical
advantages over the previous computational treatments:

\begin{itemize}
\item{No marking, sorting, or counting of constraint violations}.
\item{Application of optimality constraints is done within the
finite-state calculus}.
\item{A system of optimality constraints can be merged into a single
constraint network.}
\end{itemize}

This approach shows clearly that optimality theory is very similar to
the two older strains of finite-state phonology: classical rewrite
systems and two-level models. In optimality theory, lenient composition
plays the same role as ordinary composition in rewrite systems. The
top-down serialism of rule ordering is replaced by the left-to-right
serialism of the constraint tableau.

The new lenient composition operator has other uses beyond phonology. In
the area of syntax, Constraint Grammar (Karlsson {\em et al.}
\cite{karlsson+etal1995}) is from a formal point of view very similar to
optimality theory. Although constraint grammars so far have not been
implemented as pure finite-state systems, it is evident that the lenient
composition operator makes it possible.

\bibliographystyle{acl} 
\bibliography{/home/karttune/text/bibtex}
\end{document}